\documentclass[runningheads]{llncs}
\usepackage{graphicx}
\usepackage{amsmath}
\usepackage{cite}
\usepackage{placeins}
\usepackage{subfigure}
\usepackage{url}
\usepackage{hhline}
\usepackage{booktabs}
\usepackage{multirow}
\usepackage{amssymb}
\usepackage{cleveref}
\usepackage{color}
\usepackage[normalem]{ulem}
\usepackage[misc]{ifsym}

\newcommand\blfootnote[1]{%
  \begingroup
  \renewcommand\thefootnote{}\footnote{#1}%
  \addtocounter{footnote}{-1}%
  \endgroup
}

\begin{document}
\title{Superpixel-Guided Label Softening for Medical Image Segmentation}
	%
	%
\author{Hang Li\inst{1}\textsuperscript{*} \and
Dong Wei\inst{2}\textsuperscript{*} \and
Shilei Cao\inst{2}\textsuperscript{*} \and
Kai Ma\inst{2} \and
Liansheng Wang\inst{1}\textsuperscript{(\Letter)} \and\\
Yefeng Zheng\inst{2}}
%
%
\authorrunning{H. Li \textit{et al.}}
%
\institute{Xiamen University, Xiamen, China\\
\email{hangli@stu.xmu.edu.cn}, \email{lswang@xmu.edu.cn} \and
Tencent Jarvis Lab, Shenzhen, China\\
\email{\{donwei,eliasslcao,kylekma,yefengzheng\}@tencent.com}}
	\maketitle              
	\begin{abstract}
		Segmentation of objects of interest is one of the central tasks in medical image analysis, which is indispensable for quantitative analysis\blfootnote{\textsuperscript{*} Contributed equally.}.
		When developing machine-learning based methods for automated segmentation, manual annotations are usually used as the ground truth toward which the models learn to mimic.
		While the bulky parts of the segmentation targets are relatively easy to label, the peripheral areas are often difficult to handle due to ambiguous boundaries and the partial volume effect, \emph{etc.}, and are likely to be labeled with uncertainty.
		This uncertainty in labeling may, in turn, result in unsatisfactory performance of the trained models.
		In this paper, we propose superpixel-based label softening to tackle the above issue.
		Generated by unsupervised over-segmentation, each superpixel is expected to represent a locally homogeneous area.
If a superpixel intersects with the annotation boundary, we consider a high probability of uncertain labeling within this area.
		Driven by this intuition, we soften labels in this area based on signed distances to the annotation boundary and assign probability values within [0, 1] to them, in comparison with the original ``hard'', binary labels of either 0 or 1.
		The softened labels are then used to train the segmentation models together with the hard labels.
Experimental results on a brain MRI dataset and an optical coherence tomography dataset
demonstrate that this conceptually simple and implementation-wise easy method achieves overall superior segmentation performances to baseline and comparison methods for both 3D and 2D medical images.
	\end{abstract}
	\section{Introduction}
\begin{figure}[!t]
		\centering
		\includegraphics[width=.9\linewidth, trim={0, 17, 0, 5}, clip]{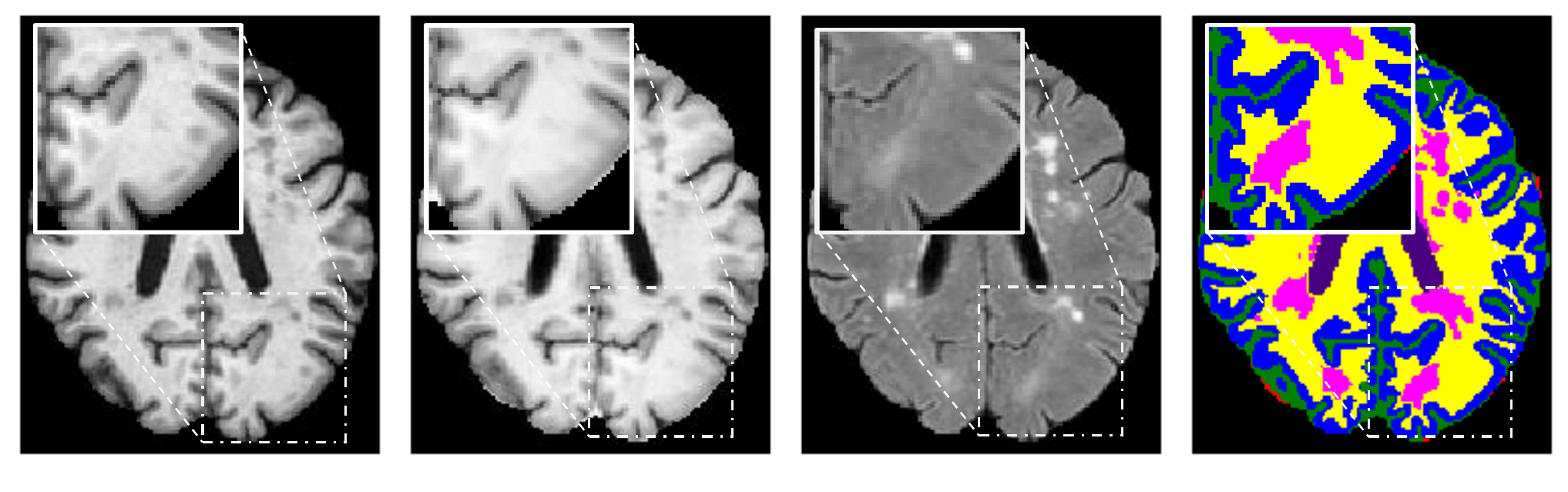}
		\caption{Illustration of ambiguous boundaries in medical images with a slice of a multi-sequence
			brain MRI.
			The first three images show the MRI sequences, and the last shows the ground truth annotation.
			As we can see, the boundaries of the tissue marked in pink are barely or even not discernible from its surroundings.
Best viewed in color.
		}\label{fig1:1}
	\end{figure}

	Segmentation of objects of interest is an important task in medical image analysis.
	Benefiting from the development of deep neural networks and the accumulation of annotated data, fully convolutional networks (FCNs) have demonstrated remarkable performances \cite{ref_unet, ref_nnunet} in this task.
	In general,
	these models assume that the ground truth is given precisely.
	However, for tasks with a large number of category labels, the peripheral areas are often difficult to annotate due to ambiguous boundaries and the partial volume effect (PVE)~\cite{ref_pve}, \emph{etc.}, and are likely to be labeled with uncertainty.
	With a limited number of data, FCNs may have difficulties in coping with such uncertainty, which in turn affects the performance.
	Taking brain MRI for example, in Fig.~\ref{fig1:1}, we show a slice of a multi-sequence MRI, in which the pink area shows barely or non-discernible boundaries from its surroundings, causing great difficulties in the manual annotation.
	
	To reduce the impact of imprecise boundary annotation, a potential solution is the label softening technique, and at this moment, we are only aware of few of them
	\cite{ref_softlabeling_lesion,ref_softlabeling_STAPLE,ref_softlaeling_unc}.
	Based on the anatomical knowledge that the lesion-surrounding pixels may also include some lesion level information, Kats \textit{et al.} \cite{ref_softlabeling_lesion} employed 3D morphological dilation to expand the binary mask of multiple sclerosis (MS) lesions and assigned a fixed pseudo probability to all pixels within the expanded region, such that these pixels can also contribute to the learning of MS lesions.
	Despite the improved Dice similarity coefficient in the experiments, the inherent contextual information of images was not utilized when determining the extent of dilation or exact value of the fixed pseudo probability.
	To account for uncertainties in the ground truth segmentation of atherosclerotic plaque in the carotid artery, Engelen \textit{et al.} \cite{ref_softlaeling_unc} proposed to blur the ground truth mask with a Gaussian filter for label softening.
	One limitation of this work was that, similar to \cite{ref_softlabeling_lesion}, the creation of the soft labels was only based on the ground truth while ignoring the descriptive contextual information in the image.
	From another perspective, soft labels can also be obtained by fusing multiple manual annotations, e.g., in \cite{ref_softlabeling_STAPLE} masks of MS lesions produced by different experts were fused using a soft version of the STAPLE algorithm \cite{warfield2004simultaneous}.
	However, obtaining multiple segmentation annotations for medical images can be practically difficult.
An alternative to label softening is the label smoothing technique~\cite{szegedy2016rethinking,ouyang2019weakly} which assumes a uniform prior distribution over labels;
yet again, this technique did not take the image context into consideration, either.
	
	In this paper, we propose a new label softening method driven by the image contextual information, for improving segmentation performance especially near the boundaries of different categories.
	Specifically, we employ the concept of superpixels \cite{ref_superpixel_paper} for the utilization of local contextual information.
	Via unsupervised over-segmentation, the superpixels group original image pixels into locally homogeneous blocks, which can be considered as meaningful atomic regions of the image.
	Conceptually, if the scale of superpixel is appropriate, pixels within the same superpixel block can be assumed belonging to the same category.
	Based on this assumption, if a superpixel intersects with the annotation boundary of the ground truth, we consider a high probability of uncertain labeling within the area prescribed by this superpixel.
	Driven by this intuition, we soften labels in this area based on the signed distance to the annotation boundary, producing probability values spanning the full range of [0, 1]---in contrast to the original ``hard'' binary labels of either 0 or 1.
	The softened labels can then be used to train the segmentation models.
	We evaluate the proposed approach on two publicly available datasets: the Grand Challenge on MR Brain Segmentation at MICCAI 2018 (MRBrainS18) \cite{ref_mrbrain_url} dataset and an optical coherence tomography (OCT) image~\cite{dme_dataset} dataset.
The experimental results verify the effectiveness of our approach.
	
	\begin{figure}[!t]
		\centering
		\includegraphics[height=.65\linewidth, trim={0, 17, 0, 10}, clip]{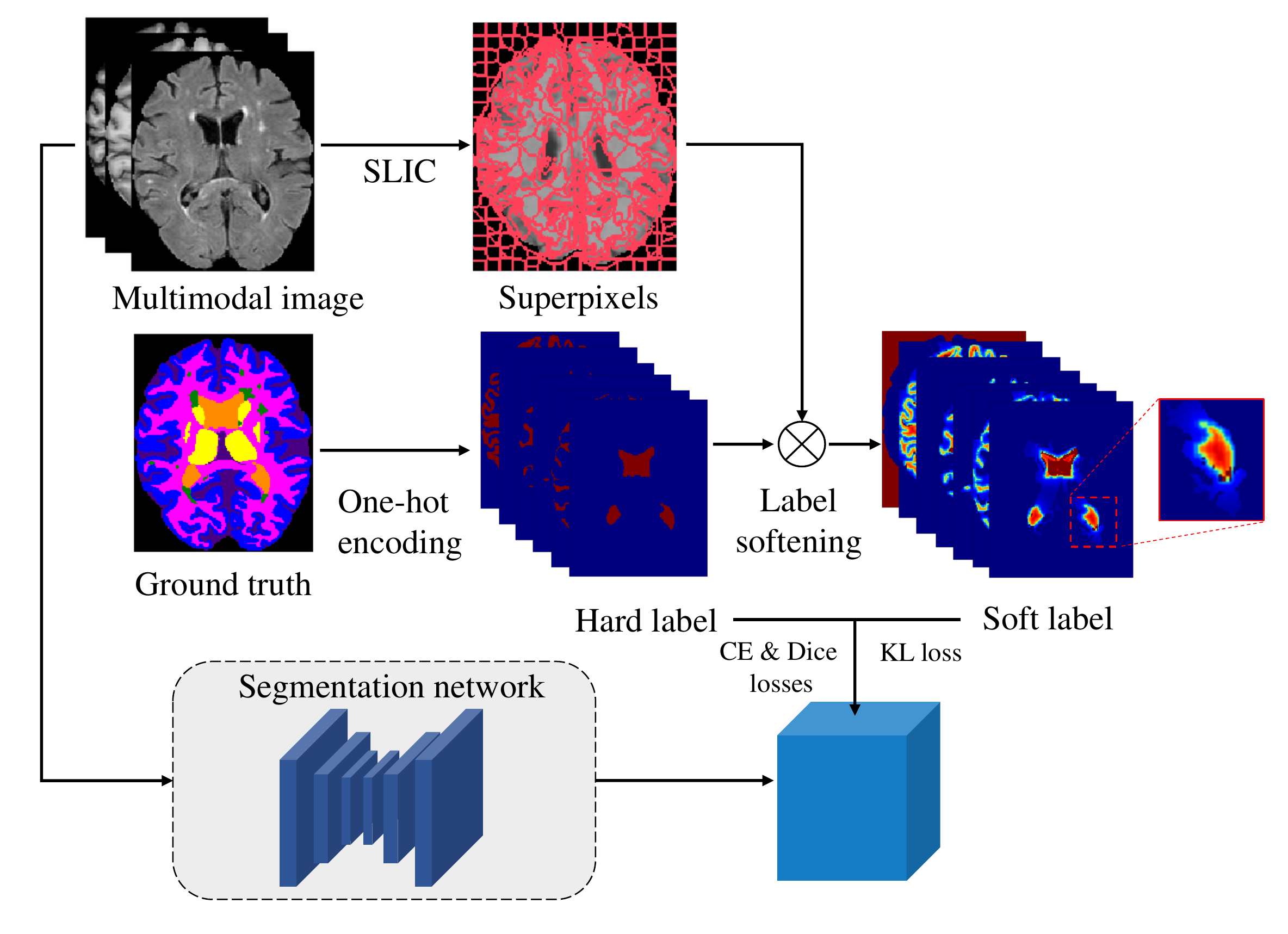}
		\caption{Pipeline of our proposed method.
		}\label{fig:arch}
	\end{figure}
	
	\section{Method}
	The pipeline of our method is illustrated in Fig. \ref{fig:arch}.
	We employ the SLIC algorithm \cite{ref_superpixel_paper} to produce superpixels, meanwhile converting the ground truth annotation to multiple one-hot label maps (the ``hard'' labels).
	Soft labels are obtained by exploiting the relations between the superpixels and hard label maps (the cross symbol $\bigotimes$ in Fig.~\ref{fig:arch}).
	Then, the soft and hard labels are used jointly to supervise the training of the segmentation network.
	
	\subsubsection{Superpixel-guided Region of Softening}
\begin{figure}[!t]
		\centering
		\includegraphics[height=.2\linewidth, trim={0, 5, 0, 2}, clip]{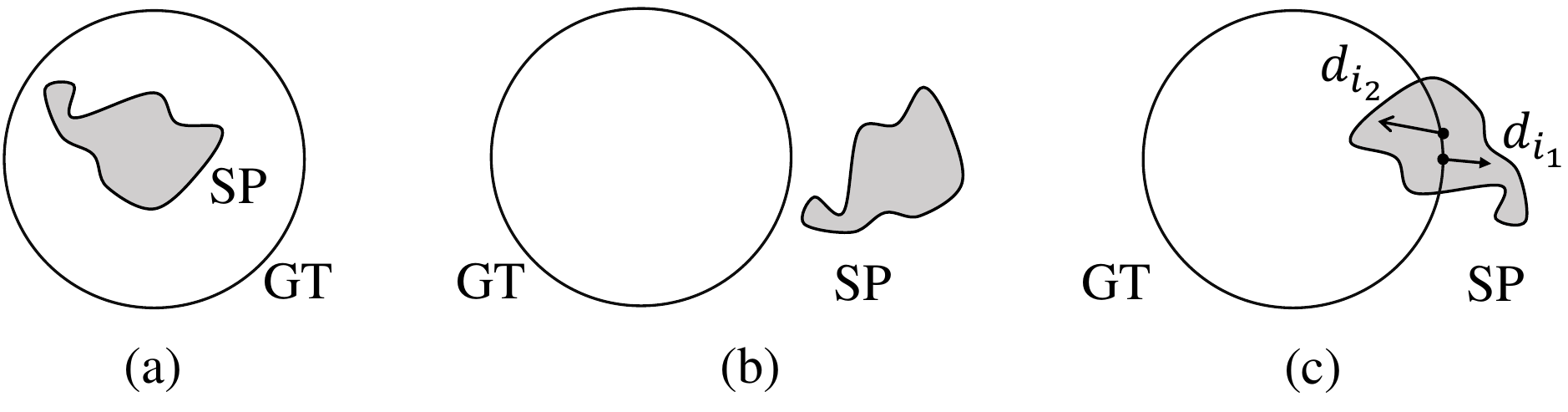}
		\caption{Illustration of three types of relations between the foreground region in a binary ground truth label map (GT) and a superpixel block (SP).
			(a) SP is inside GT; (b) SP is outside GT; (c) SP intersects with boundaries of GT.
			We identity the region enclosed by the SP in the third case for label softening, based on the signed distances to GT boundaries.}
		\label{fig_loc:loc}
	\end{figure}

	Our purpose is to model the uncertainty near the boundaries of categories in the manual annotation for improving model performance and robustness.
	For this purpose, we propose to exploit the relations between superpixels and the ground truth annotation to produce soft labels.
	Specifically, we identify three types of relations between a superpixel and the foreground region in a one-hot ground truth label map (Fig. \ref{fig_loc:loc}):
	(a) the superpixel is inside the region,
	(b) the superpixel is outside the region, and
	(c) the superpixel intersects with the region boundary.
	As the superpixel algorithms \cite{ref_superpixel_paper} group pixels into locally homogeneous pixel blocks, pixels within the same superpixel can be assumed to belong to the same category given that superpixels are set to a proper size.
	Based on this assumption, it is most likely for uncertain annotations to happen in the last case, where the ground truth annotation indicates different labels for pixels inside the same superpixel block.
	Therefore, our label softening works exclusively in this case.
	
	Formally, let us denote an image by $x \in \mathbb{R}^{W \times H}$, where $W$ and $H$ are the width and height, respectively.
	(Without loss of generalization, $x$ can also be a 3D image $x \in \mathbb{R}^{W \times H \times T}$, where $T$ is the number of slices, and our method still applies.)
	Then, its corresponding ground truth annotation can be denoted by a set of one-hot label maps: $Y=\{y^c | y^c\in \mathbb{R}^{W \times H}\}_{c=1}^C$, where $C$ is the number of categories, and $y^c$ is the binary label map for category $c$, in which any pixel $y^c_i\in\{0, 1\}$, where $i\in\{1,\ldots,N\}$ is the pixel index, and $N$ is the total number of pixels;
	besides, we denote the foreground area in $y^c$ by $\phi^c$.
	We can generate superpixel blocks $S(x)=\{s^{(j)}\}_{j=1}^M$ for $x$ using an over-segmentation algorithm, where $M$ is the total number of superpixels.
	In this paper, we adopt SLIC \cite{ref_superpixel_paper} as our superpixel-generating algorithm, which is known for computational efficiency and quality of the generated superpixels.
	We denote the set of soft label maps to be generated by $Q_c=\{q^c | q^c \in \mathbb{R}^{W \times H}\}$;
	note that $q^c_i\in[0,1]$ is a continuous value, in contrast with the binaries in $y^c$.
	As shown in Fig.~\ref{fig_loc:loc}, the relations between any $\phi^c$ and $s^{(j)}$ can be classified into three categories: (a) $s^{(j)}$ is inside $\phi^c$; (b) $s^{(j)}$ is outside $\phi^c$; and (c) $s^{(j)}$ intersects with boundaries of $\phi^c$.
	For the first two cases, we use the original values of $y_i^c$ in the corresponding locations in $q^c$.
	Whereas as for the third case, we employ label softening strategies to assign a soft label $q^c_i$ to each pixel $i$ based on its distance to boundaries of $\phi^c$, which is described below.
	
	\subsubsection{Soft Labeling with Signed Distance Function}
	Assume a superpixel block $s$ intersects with the boundaries of a foreground $\phi$ (for simplicity, the superscripts can be safely omitted here without confusion).
	For a pixel $s_i$ in $s$, the absolute value of the distance $d_i$ from $s_i$ to $\phi$ is defined as the minimum distance among all the distances from $s_i$ to all pixels on the boundaries of $\phi$.
	We define $d_i > 0$ if $s_i$ is inside $\phi$, and $d_i \leq 0$ otherwise.
	As aforementioned, in the case of a superpixel block intersecting with the boundaries of $\phi$, we need to assign each pixel in this block a pseudo-probability as its soft label according to its distance to $\phi$.
	The pseudo-probability should be set to $0.5$ for a pixel right on the boundary (i.e. $d_i=0$), gradually approach 1 as $d_i$ increases, and gradually approach 0 otherwise.
	Thus, we define the distance-to-probability conversion function as
	\begin{equation}\label{eq:d_to_p}
	q_i=f_\mathrm{dist}(d_i)=\frac{1}{2}\left(\frac{d_i}{1+|d_i|}+1\right),
	\end{equation}
	where $q_i\in[0,1]$ is the obtained soft label for pixel $i$.
	
	\subsubsection{Model Training with Soft and Hard Labels}
	We adopt the Kullback-Leibler (KL) divergence loss~\cite{kullback1951information} to supervise model training with our soft labels:
	\begin{equation} \label{kl_loss_func}
	\mathcal{L}_\mathrm{KL} = \frac{1}{N}{\sum}_{i=1}^N {\sum}_{c=1}^C q_{i}^c \log \left(q_{i}^c/p_{i}^{c}\right),
	\end{equation}
	where $p_{i}^c$ is the predicted probability of the $i$-th pixel belonging to the class $c$, and $q_{i}^c$ is the corresponding soft label defined with Equation~(\ref{eq:d_to_p}).
	Along with $\mathcal{L}_\mathrm{KL}$, we also adopt the commonly used Dice loss $\mathcal{L}_\mathrm{Dice}$~\cite{ref_dice_loss} and cross-entropy (CE) loss $\mathcal{L}_\mathrm{CE}$ for medical image segmentation.
	Specifically,
the CE loss is defined as:
	\begin{equation} \label{ce_loss}
	\mathcal{L}_\mathrm{CE} = -\frac{1}{N}{\sum}_{i=1}^N{\sum}_{c=1}^C w_{c}y_{i}^c\log (p_i^c),
	\end{equation}
	where $w_c$ is the weight for class $c$.
	When $w_c=1$ for all classes, Equation~(\ref{ce_loss}) is the standard CE loss.
In addition, $w_c$ can be set to class-specific weights to counteract the impact of class imbalance~\cite{paszke2016enet}: $w_c= 1 / \log(1.02 + {\sum}_{i=1}^N y_i^c/N)$,
and we refer to this version of the CE loss as weighted CE (WCE) loss.
	The final loss is defined as a weighted sum of the three losses:
	$\mathcal{L}  = \mathcal{L}_\mathrm{CE} + \alpha \mathcal{L}_\mathrm{Dice} + \beta \mathcal{L}_\mathrm{KL}$,
	where $\alpha$ and $\beta$ are hyperparameters to balance the three losses.
	We follow the setting in nnU-Net\cite{ref_nnunet} to set $\alpha=1.0$, and explore the proper value of $\beta$ in our experiments, since it controls the relative contribution of our newly proposed soft labels which are of interest.
	
	\section{Experiments}
	\subsubsection{Datasets}
	To verify the effectiveness of our method on both 2D and 3D medical image segmentation, we use datasets of both types for experiments.
	The MRBrainS18 dataset \cite{ref_mrbrain_url} provides
	seven 3T multi-sequence (T1-weighted, T1-weighted inversion recovery, and T2-FLAIR) brain MRI scans
	with the following 11 ground truth labels: 0-background, 1-cortical gray matter, 2-basal ganglia, 3-white matter, 4-white matter lesions, 5-cerebrospinal fluid in the extracerebral space, 6-ventricles, 7-cerebellum, 8-brain stem, 9-infarction and 10-other, among which labels 9 and 10 were officially excluded from the evaluation and we follow this setting.
	We randomly
	choose five scans for training and use the rest for evaluation.
	For preprocessing, the scans are preprocessed by skull stripping, nonzero cropping, resampling, and data normalization.
	The other dataset \cite{dme_dataset} includes OCT images with diabetic macular edema (the OCT-DME dataset) for the segmentation of retinal layers and fluid regions.
	It contains 110 2D B-scan images
	from 10 patients.
	Eight retinal layers and fluid regions are annotated. We use the first five subjects for training and the last five subjects for evaluation (each set has 55 B-scans).
	Since the image quality of this dataset is poor, we firstly employ a denoising convolutional neural networks (DnCNN) \cite{ref_DnCNN} to reduce image noise and improve the visibility of anatomical structures.
To reduce memory usage, we follow He \emph{et al.}~\cite{he2019fully} to flatten a retinal B-scan image to the estimated Bruch’s membrane (BM) using an intensity gradient method \cite{lang2013retinal} and crop the retina part out.
	
	\subsubsection{Experimental Setting and Implementation}
	For the experiments on each dataset, we first establish a baseline, which is trained without the soft labels.
	Then, we re-implement the Gaussian blur based label softening method~\cite{ref_softlaeling_unc}, in which the value of $\sigma$ is empirically selected, for a comparison with our proposed method.
	Considering the class imbalance in both datasets, we present results using the standard CE and WCE losses for all methods.
	We notice that the Dice loss adversely affects the performance on the OCT-DME dataset, therefore those results are not reported.
	We use overlap-based, volume-based, and distance-based mean metrics \cite{taha2015metrics}, including: Dice coefficient score, volumetric similarity (VS), 95\textsuperscript{th} percentile Hausdorff distance (HD95), average surface distance (ASD), and average symmetric surface distance (ASSD) for a comprehensive evaluation of the methods.
	We employ a 2D U-Net \cite{ref_unet} segmentation model (with the Xception \cite{ref_xception} encoder) for the OCT-DME dataset, and a 3D U-Net \cite{ref_nnunet} model for the MRBrainS18 dataset (patch-based training and sliding window test tricks \cite{jin2019accurate} are employed in the implementation).
	All experiments are conducted with the PyTorch framework \cite{pytorch} on a standard PC with an NVIDIA GTX 1080Ti GPU.
	The Adam optimizer\cite{ref_adam} is adopted with a learning rate of $3\times 10^{-4}$ and a weight decay of $10^{-5}$.
	The learning rate is halved if the validation performance does not improve for 20 consecutive epochs. The batch size is fixed to 2 for the MRBrainS18 dataset, and 16 for the OCT-DME dataset.
	
	
	
\subsubsection{Results}
The quantitative evaluation results are summarized in Table \ref{mrbrain_result_table} and Table \ref{oct_result_table} for the MRBrainS18 and OCT-DME datasets, respectively.
(Example segmentation results on both datasets are provided in the supplementary material.)
As expected, the weighted CE loss produces better results than the standard CE loss for most evaluation metrics on both datasets.
We note that the Gaussian blur based label softening \cite{ref_softlaeling_unc} does not improve upon the baselines either with the CE or WCE loss, but only obtains results comparable to those of the baselines.
The reason might be that this method indiscriminately softens all boundary-surrounding pixels with a fixed standard deviation without considering the actual image context, which may potentially harm the segmentation near originally precisely annotated boundaries.
In contrast, our proposed method consistently improves all metrics when using the generated soft labels with the WCE loss.
In fact, with this combination of losses, our method achieves the best performances for all evaluation metrics.
It is also worth mentioning that, although our method is motivated by improving segmentation near category boundaries, it also improves the overlap-based evaluation metrics (Dice) by a noticeable extent on the OCT-DME dataset.
These results verify the effectiveness of our method in improving segmentation performance, by modeling uncertainty in manual labeling with the interaction between superpixels and ground truth annotations.
	
	\begin{table}[!t]
		\newcommand{\tabincell}[2]{\begin{tabular}{@{}#1@{}}#2\end{tabular}}
		\caption{Evaluation results on the MRBrainS18 dataset \cite{ref_mrbrain_url}.
			The KL divergence loss is used by our method for model training with our soft labels.}\label{mrbrain_result_table}
		\scalebox{1}{\centering
        \setlength{\tabcolsep}{.2mm}{
		\begin{tabular}{l|l|c|c|c|c|c}
			\hline
			
			Method        & Losses            &  Dice (\%)$\uparrow$  & VS (\%)$\uparrow$ & HD95 (mm)$\downarrow$ &
			ASD (mm)$\downarrow$  & ASSD (mm)$\downarrow$ \\
			\hhline{=======}
			\multirow{2}*{Baseline} & CE+Dice&      85.47      &                      96.53                       &      3.5287      &      0.9290      &      0.7722      \\
			&WCE+Dice&      85.56      &                      96.55                       &      3.5116      &      0.8554      &      0.7371      \\
			\hline
			Engelen&
			CE+Dice& 85.16 & 95.44& 3.6396 & 0.9551 & 0.8004 \\
			 \textit{et al.} \cite{ref_softlaeling_unc} &WCE+Dice& 85.47 & 95.58& 3.5720 & 0.8432 & 0.7502 \\
			\hline
			\multirow{2}*{Ours}&KL+CE+Dice&      85.26      &                      93.95                       &      3.4684      &      0.9460      &      0.8061      \\

			&KL+WCE+Dice & \bfseries 85.63 &                 \bfseries 96.60                  &  \bfseries    3.1194      & \bfseries 0.8146 & \bfseries 0.7153 \\ \hline
		\end{tabular}}}
	\end{table}
	
	
\begin{table}[!t]
		\newcommand{\tabincell}[2]{\begin{tabular}{@{}#1@{}}#2\end{tabular}}
		\caption{Evaluation results on the OCT-DME dataset \cite{dme_dataset}.
			The KL divergence loss is used by our method for model training with our soft labels.}\label{oct_result_table}
		\scalebox{.97}{\centering
        \setlength{\tabcolsep}{1.05mm}{
		\begin{tabular}{l|l|c|c|c|c|c}
			\hline
			Method     & Losses        &  Dice (\%)$\uparrow$ &  VS (\%)$\uparrow$ & HD95 (mm)$\downarrow$ & ASD (mm)$\downarrow$ & ASSD (mm)$\downarrow$  \\
			\hhline{=======}
			\multirow{2}*{Baseline} &CE &    $ 82.50 $   & $ 96.52 $   &      $3.22$    &  $1.075$ & $1.075$   \\
			&WCE      & $82.82 $ &  $96.44$ & $3.27$  &  $1.082$& $1.087$     \\
			\hline
				Engelen &CE & $82.69$ & $96.38$ & $ 3.24$  & $1.092 $  & $1.092$ \\
			\textit{et al.} \cite{ref_softlaeling_unc}&WCE& $82.94 $ & $96.36 $ & $3.27 $  & $1.080 $ & $1.094 $ \\
			\hline
			\multirow{2}*{Ours}&KL+CE & $82.72$ & $96.41 $ & $3.25$ & $1.081 $ & $1.087$ \\
			&KL+WCE&    \boldmath{$83.94 $}  & \boldmath $96.73  $  & \boldmath $3.17$ & \boldmath $1.058 $& \boldmath $1.066 $  \\ \hline
		\end{tabular}}}
	\end{table}
	
	\subsubsection{Ablation Study on Number of Superpixels}
\begin{figure}[!t]
		\centering
		\includegraphics[width=.95\linewidth, trim={0, 7, 0, 10}, clip]{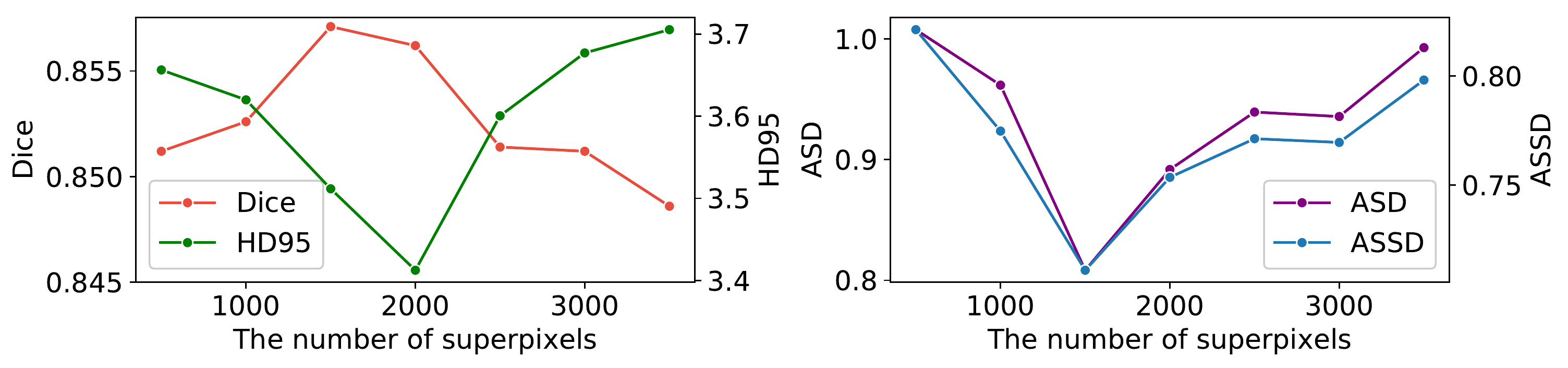}
		\caption{Performances of our method with different numbers of superpixels on the MRBrainS18 dataset \cite{ref_mrbrain_url}. The HD95, ASD and ASSD are in mm.
Best viewed in color.}
		\label{fig:mrbrainremake}
	\end{figure}

	The proper scale of the superpixels is crucial for our proposed method, as superpixels of different sizes may describe different levels of image characteristics, and thus may interact differently with the ground truth annotation.
	Since in the SLIC \cite{ref_superpixel_paper} algorithm, the size of superpixels is controlled by the total number of generated superpixel blocks, we conduct experiments to study how the number of superpixels influences the performance on the MRBrainS18 dataset.
	In Fig. \ref{fig:mrbrainremake}, we show performances of our method with different numbers of superpixels ranged from 500 to 3500 with a sampling interval of 500.
	As we can see, as the number of superpixels increases, the performance first increases due to the more image details incorporated, and then decreases after reaching the peak.
	This is in line with our intuition, since the assumption that pixels within the same superpixel belong to the same category can hold only if the scale of superpixels is appropriate.
	Large superpixels can produce flawed soft labels.
	In contrast, as the number of superpixels grows and their sizes shrink, soft labels will degenerate into hard labels, which does not provide additional information.
	
	\begin{figure}[t]
		\centering
		\includegraphics[width=.95\linewidth, trim={0, 9, 0, 6}, clip]{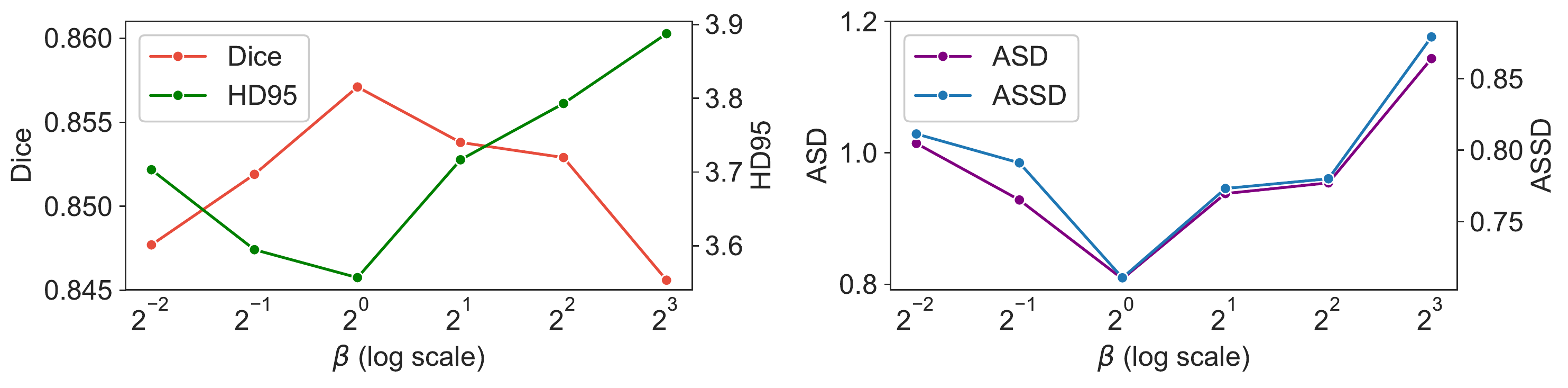}
		\caption{Performances of our method using different values of $\beta$ on the MRBrainS18 dataset \cite{ref_mrbrain_url}.
			The HD95, ASD and ASSD are in mm.
Best viewed in color.}
		\label{fig:mrbrainkl}
	\end{figure}
	
	\subsubsection{Ablation Study on Weight of Soft Label Loss}
	The weight $\beta$
controls the contribution of the soft labels in training.
	To explore the influence of the soft label loss, we conduct a study on the MRBrainS18 dataset to compare the performance of our method with different values of $\beta$.
	We set $\beta$ to $1/4, 1/2, 1, 2, 4$, and $8$.
	The mean Dice, HD95, ASD, and ASSD of our proposed method with these values of $\beta$ are shown in Fig. \ref{fig:mrbrainkl}.
	Note that the $x$-axis uses a log scale since values of $\beta$ differ by orders of magnitude.
	Improvements in performance can be observed when $\beta$ increases from $1/4$ to 1.
	When $\beta$ continues to increase, however, the segmentation performances start to drop.
	This indicates that the soft labels are helpful to segmentation, although giving too much emphasis to them may decrease the generalization ability of the segmentation model.

	\section{Conclusion}
	In this paper, we presented a new label softening method that was simple yet effective in improving segmentation performance, especially near the boundaries of different categories.
	The proposed method first employed an over-segmentation algorithm to group image pixels into locally homogeneous blocks called superpixels.
	Then, the superpixel blocks intersecting with the category boundaries in the ground truth were identified for label softening,
	and a signed distance function was employed to convert the pixel-to-boundary distances to soft labels within $[0,1]$ for
pixels inside these blocks.
	The soft labels were subsequently used to train a segmentation network.
	Experimental results on both 2D and 3D medical images demonstrated the effectiveness of this simple approach in improving segmentation performance.

\subsubsection{Acknowledgments.} This work was funded by the National Natural Science Foundation of China (Grant No. 61671399), Fundamental Research Funds for the Central Universities (Grant No. 20720190012), Key Area Research and Development Program of Guangdong Province, China (No. 2018B010111001), National Key Research and Development Project (2018YFC2000702), and Science and Technology Program of Shenzhen, China (No. ZDSYS201802021814180).

	\FloatBarrier

	%
	%
		\bibliographystyle{splncs04}
		\bibliography{paper1230}
	
	\newpage
	\begin{center}
		\textbf{\large Supplementary Material: Superpixel-Guided Label Softening for Medical Image Segmentation}
	\end{center}
	\setcounter{equation}{0}
	\setcounter{figure}{0}
	\setcounter{table}{0}
	\setcounter{page}{1}
	\makeatletter
	\renewcommand{\theequation}{S\arabic{equation}}
	\renewcommand{\thefigure}{S\arabic{figure}}
	\renewcommand{\theproposition}{S\arabic{proposition}}
	\begin{figure}[h]
		\centering
	\includegraphics[width=.95\linewidth]{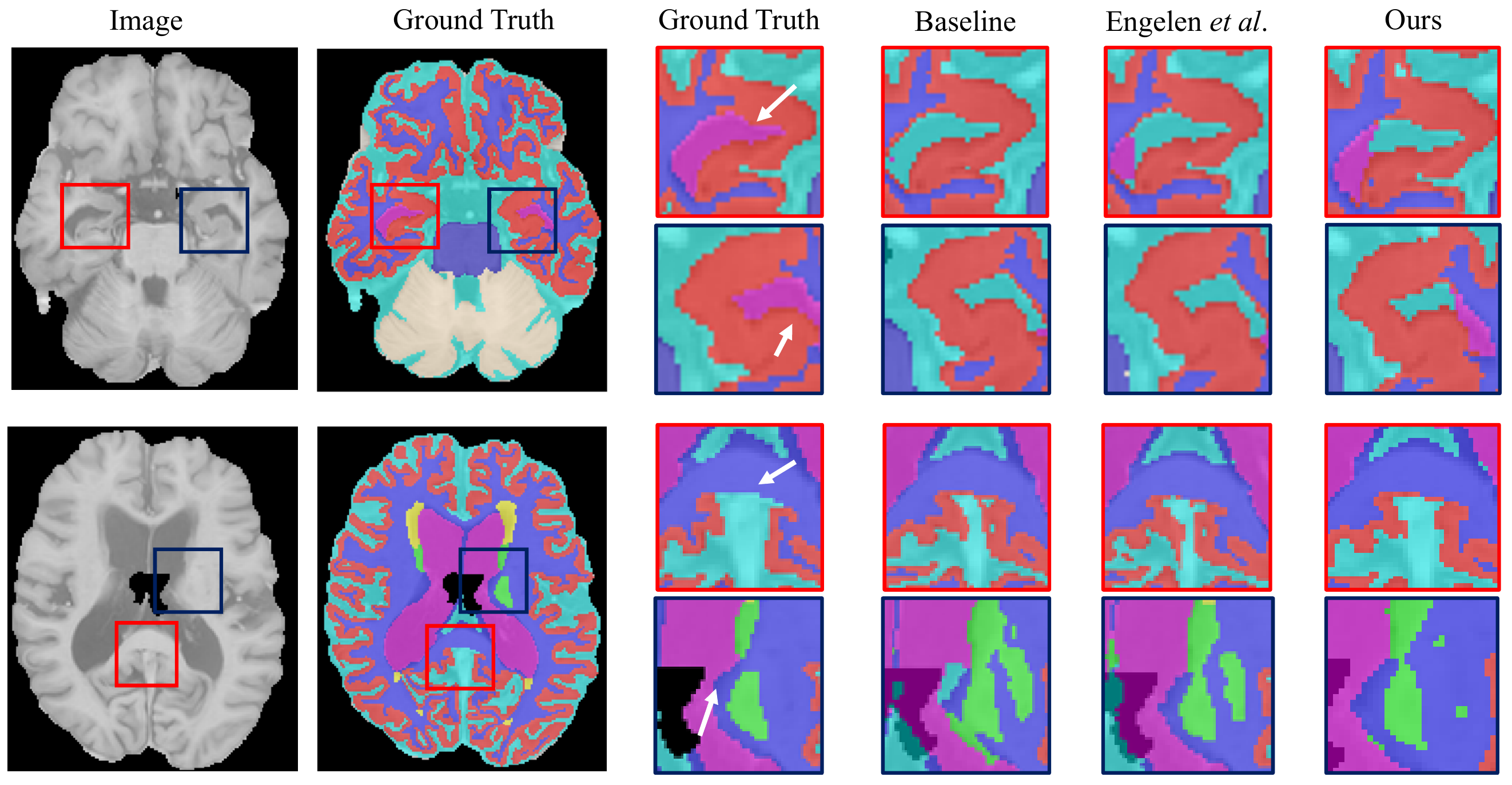}
		\caption{Example segmentation results on the MRBrainS18 dataset~\cite{ref_mrbrain_url}.
Best viewed in color.}
		\label{fig:visiualmrbrains}
	\end{figure}

	\begin{figure}[h]
		\centering
		\includegraphics[width=.95\linewidth]{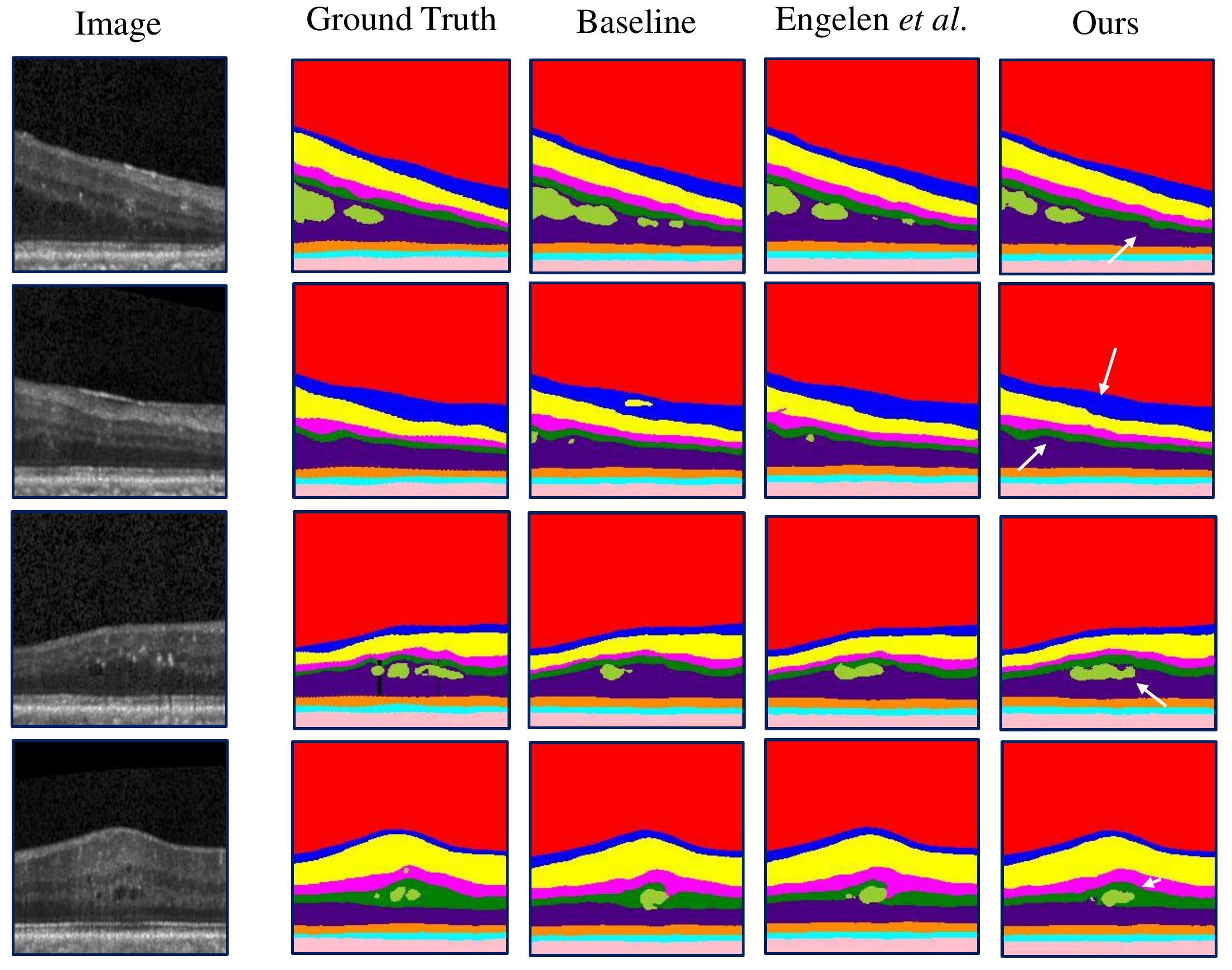}
		\caption{Example segmentation results on the OCT-DME dataset~\cite{dme_dataset}.
Best viewed in color.}
		\label{fig:visiualoct}
	\end{figure}

\end{document}